\documentclass[letterpaper, 10 pt, conference]{ieeeconf}  

\IEEEoverridecommandlockouts                              

\usepackage{graphicx} 
\usepackage{amsmath,amssymb}
\usepackage{multirow}
\usepackage{booktabs}

\newtheorem{definition}{Definition}
\newtheorem{assumption}{Assumption}

\title{\LARGE \bf
Task-Space Motion-Force Nonlinear Model Predictive Control for Precise and Compliant Control under Uncertainties
}

\author{Quang-Nam Nguyen$^{1}$ and Ludovic Righetti$^{1,2}$
\thanks{$^{1}$Machines in Motion Laboratory, Center for Robotics and Embodied 
        Intelligence, New York University (NYU), USA. 
        {\tt\small namnguyen@nyu.edu}
}
\thanks{$^{2}$Artificial and Natural Intelligence Toulouse Institute (ANITI), France.}%
}

\begin{document}

\maketitle
\thispagestyle{empty}
\pagestyle{empty}

\begin{abstract}
In Nonlinear Model Predictive Control (NMPC), the control law is implicitly determined by the solution of an Optimal Control Problem (OCP). 
Consequently, OCP design is often tailored to specific platforms and tasks, demanding extensive expert knowledge and tuning effort. 
To address this, we propose a unified NMPC framework that directly plans and controls the robot joints subject to task-space motion and force requirements and is capable of seamlessly generalizing across multiple robotic platforms and diverse tasks. 
Furthermore, optimization-based methods often treat interactions as rigid contacts, making the torque solution sensitive to the modeling errors of the robot dynamics and the environment. 
We study the intrinsic task-space impedance of our proposed controller, its partial contraction for differential stability and differential passivity under unmodeled interactions.
As a result, our framework enables closing the force-feedback loop in the high-level planning layer to achieve precise motion-force tracking performance, addressing the long-standing challenge of force-feedback NMPC.
Finally, we validate our framework on physical hardware, demonstrating hybrid motion-force control on a redundant manipulator and compliant locomotion on a quadruped, while preserving core NMPC advantages such as hard constraint satisfaction, optimal redundancy exploitation, and real-time planning under unforeseen interactions.
\end{abstract}

\section{INTRODUCTION}

Recent progress in model-based optimization has improved the ability to solve real-time whole-body Optimal Control Problems (OCP) for Nonlinear Model-Predictive Control (NMPC), thanks to the cumulative efforts of dynamics algorithms \cite{carpentier2019pinocchio}, versatile OCP formulation \cite{mastalli2020crocoddyl}, and efficient solvers \cite{jordana2025structure}.
In general, a discrete OCP with horizon $N$ aiming to minimize a cost function under dynamics constraints and inequality constraints can be written as:
\begin{subequations} \label{eq:ocp_discrete}
\begin{align}
    &\min_{\mathbf{x,u}} \sum_{k=0}^{N-1} l_k(x_k,u_k) + l_N(x_N) \label{cost}\\
    \text{s.t.} \quad &x_{k+1} = f_k(x_k,u_k) \quad \forall k \in [0,N-1]\\
    &h_k(x_k,u_k) \leq 0 \label{ineq_const} \quad \forall k \in [0,N]
\end{align}
\end{subequations}
where the constraints (\ref{eq:ocp_discrete}c) may include joint state and torque limits, task-space collision avoidance, friction cones, etc.

Such approaches are interesting as they offer, in theory, the ability to adapt in real-time to any types of environment while ensuring hard constraint satisfaction when necessary.
In practice, however, OCP has been mostly used in offline trajectory optimization due to the computational burden of long-horizon OCP with complex costs and constraints designed to solve a given task.
To realize the motion on real robots, the optimal trajectory is then used as reference for faster tracking controllers such as joint-space linear/nonlinear MPC \cite{bishop2025surprising} or reinforcement learning (RL) tracking policies \cite{taouil2026motiondisco}.
Recent efforts have made whole-body NMPC possible by carefully tailoring the OCP to specific platforms and tasks \cite{khazoom2024tailoring} or making strong assumptions about contacts, either via analytical model \cite{kleff2022introducing} or in simulation \cite{zhang2025whole}, rendering these approaches often brittle to unexpected interactions.
These reasons can explain why the community tends to prefer policy learning methods (e.g. RL) \cite{cao2024efficient, taouil2026motiondisco} which remove all the issues of real-time decision making.
However, such approaches might not be suitable for all contexts, for example when hard constraint satisfaction guarantees are dictated by the application, or when the environment model is inaccurate or unavailable, leading to large sim-to-real gap.
Furthermore, incorporating force/torque sensors to achieve motion-force tracking performance of state-of-the-art force control techniques has been a challenge for both optimization and learning methods \cite{kleff2024force}.

\begin{figure}
    \centering
    \includegraphics[width=0.9\linewidth]{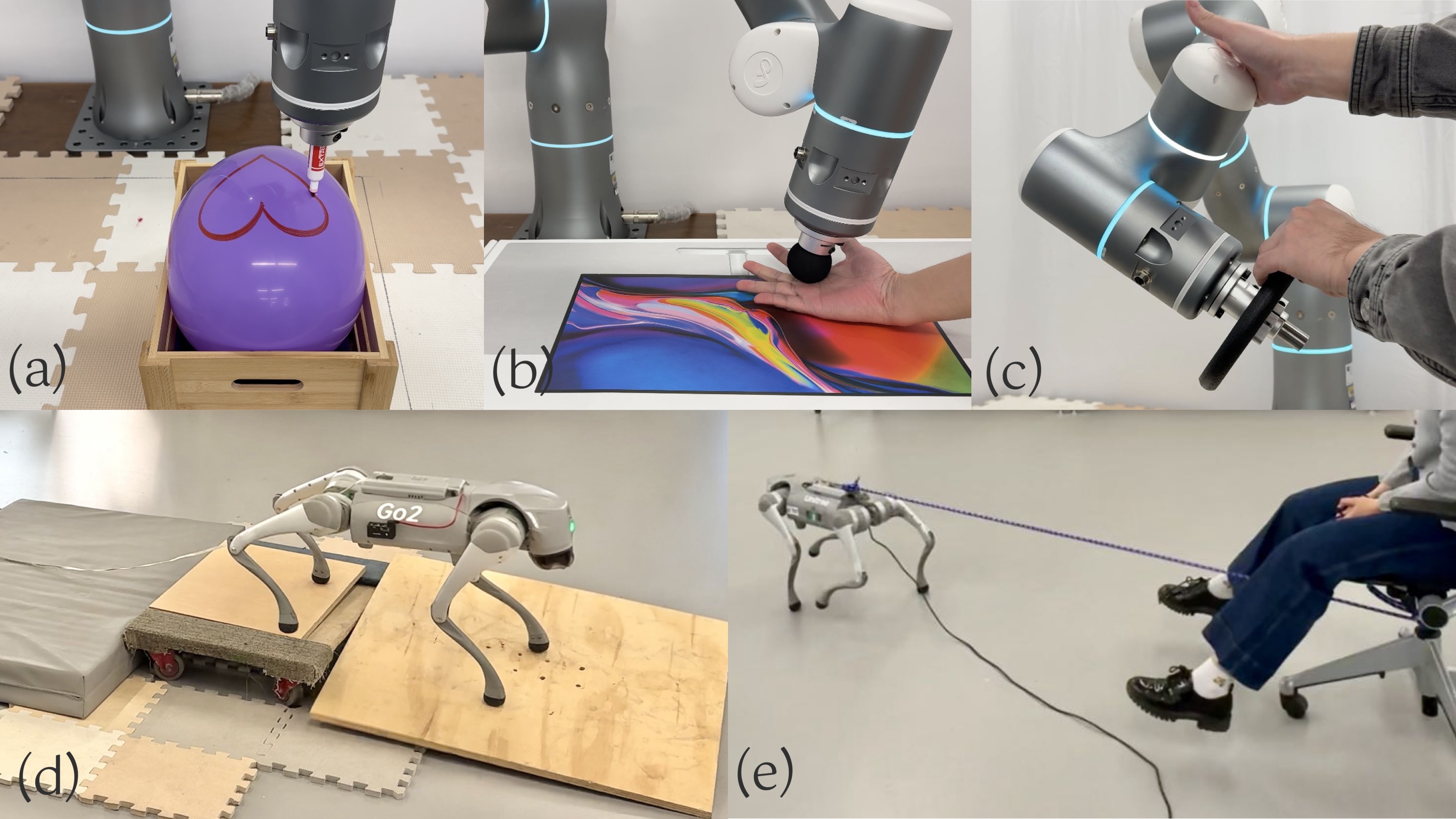}
    \caption{Our force feedback IA-NMPC performs safe and precise motion-force control tasks: 
    (a) drawing on a deformable surface, 
    (b) massaging, and 
    (c) $SE(3)$ manipulation of an unknown heavy object.
    Without force feedback, our I-NMPC creates robust, compliant behaviors on a quadruped:
    (d) walking through unknown terrains, and
    (e) pulling human on a chair.}
    \label{fig:highlights}
\end{figure}

In this paper, we propose a task-space NMPC framework to generate in real-time robot motion that guarantees passive impedance in the task space for various robots.
Indeed, a diverse set of robotic tasks can be achieved with platform-agnostic task-space targets, then a generic OCP featuring end-effector reaching costs can generalize to multiple robot platforms.
Moreover, based on theoretical analysis, we remove unnecessary rigid contact assumptions and instead choose appropriate end-effector cost features to regulate the robot's task-space impedance and its overall movement in a generic manner while increasing its robustness to disturbances and model errors. 
As a surprising result, this framework also enables us to utilize a simple and modular way to close the force-feedback loop in NMPC through admittance control.
We validate our approach for multiple motion-force control tasks on a 7-DOF manipulator and achieve the same tracking precision as state-of-the-art force control methods while preserving the additional advantages of NMPC such as optimal redundancy resolution and real-time predictive planning under hard constraint satisfaction and unmodeled human-robot interactions.
We also run NMPC on a real quadruped to stress-test the robustness to unknown environments and modeling/sensory uncertainties, while achieving customizable compliance at the robot's feet and center of mass (COM).
We open-source our framework which seamlessly controls diverse types of robots for different tasks, utilizing Pinocchio for our dynamics model and Mujoco for simulations to mimic the model-to-real gap.

\subsection{Related work}
\subsubsection{Control by optimization}
In recent years, the robotics community has largely moved from closed-form methods such as operational-space control \cite{khatib1987unified}, inverse dynamics control \cite{righetti2011inverse,peters2005unifying}, to optimization-based methods aiming to resolve an OCP, either online through MPC or offline with RL.
Optimizing a cost function under constraints, i.e. an OCP, is equivalent to a control law, for example, in the Linear Quadratic Regulator (LQR).
However, it is not always possible to obtain an analytical form of the control law because the difficulty levels between solving differential equations and solving optimization directly may differ for different problems, especially when the dynamics is nonlinear.
In this paper, we would like to recover the underlying control law resulting from a generic OCP under nonlinear dynamics with simple joint velocity/torque regularization and end-effector costs, which can be used in diverse tasks such as pick-and-place, goal reaching, drawing, polishing, walking, etc.

Classical methods often prefer to decouple complex tasks into task-space planning followed by joint-space planning and control. 
Similarly, recent efforts in learning-based methods also aim to remove joint-space tele-operation data, which are expensive and hard to scale, and instead collect demonstration data in task space \cite{xu2026hommi}. 
Following this paradigm, we propose a generic NMPC architecture (see Fig. \ref{fig:control_architecture}) that is applicable to a wide range of robots and replaces Inverse Kinematics (IK), Inverse Dynamics (ID), Proportional-Derivative (PD) control with receding-horizon optimal control.

\subsubsection{Impedance and Admittance}
Numerous methods have been developed to address hybrid motion-force control, including impedance \cite{erhart2013impedance} and admittance \cite{tarbouriech2019admittance} which aim to establish a mass-spring-damper relationship for end-effector contacts but differ in whether the end-effector is force-controlled or position-controlled. 
For locomotion, COM admittance and foot damping can be modeled via an admittance control law \cite{caron2019stair} or a PD control law \cite{focchi2017high} and eventually mapped to joint state and torque using IK and ID solvers \cite{herzog2016momentum}.
The relationship between impedance and admittance is known to be complementary, and many works have proposed unifying impedance and admittance, such as mode switching \cite{ott2015hybrid} or through hardware design \cite{elwin2022human}. 
Unlike in closed-form methods which specify spring-like and damper-like constants \cite{hogan1985impedance}, we can obtain impedance stiffness that can be scaled by the weight matrices in nonlinear OCP.

\subsubsection{Force-feedback MPC}
There are two approaches in force-feedback MPC which originate in force control: direct and indirect.
Some direct force-control MPC methods may impose the rigid contact constraint on the end-effectors \cite{kleff2022introducing, jordana2024force}, which over-constrains the robot while requiring accurate contact timings and locations.
Moreover, contact forces and control torques are algebraically coupled due to the over-constrained dynamics.
Although some works have proposed breaking the algebraic loop by introducing delay \cite{kleff2022introducing} or using soft contact models \cite{kleff2024force}, these methods introduce task-specific contact assumptions, making it difficult to implement for different tasks and sensitive to actual contact state.
In comparison, we do not apply rigid contact constraints to interactions except standing foot of legged robots.
We show that the mismatch between our predictive model and the actual interaction helps characterize task-space impedance, and afford reasoning about contacts in task space instead without modifying the dynamics model.

We follow the indirect approach which has shown promising results in linear and nonlinear MPC.
For example, \cite{pankert2020perceptive} proposed to adjust reference end-effector trajectory to regulate contact forces, i.e. an admittance control scheme, in a position-controlled mobile manipulator using sequential linear quadratic MPC.
In comparison, our method uses an admittance loop to regulate impedance instead of relying only on the actual robot movements.
Recently, \cite{alharbat2025predictive} combined admittance control with NMPC for aerial robots, updating the reference trajectory using the impedance model's predictions.
In contrast, our method does not use interaction forces as an optimization variable nor in the OCP's dynamics model, keeping our formulation generic and avoiding the need for force derivative, which is commonly set to zero \cite{alharbat2025predictive} or estimated indirectly with a contact dynamics model \cite{kleff2024force}.

\section{METHODOLOGY OVERVIEW}

\begin{figure}
    \centering
    \includegraphics[width=0.6\linewidth]{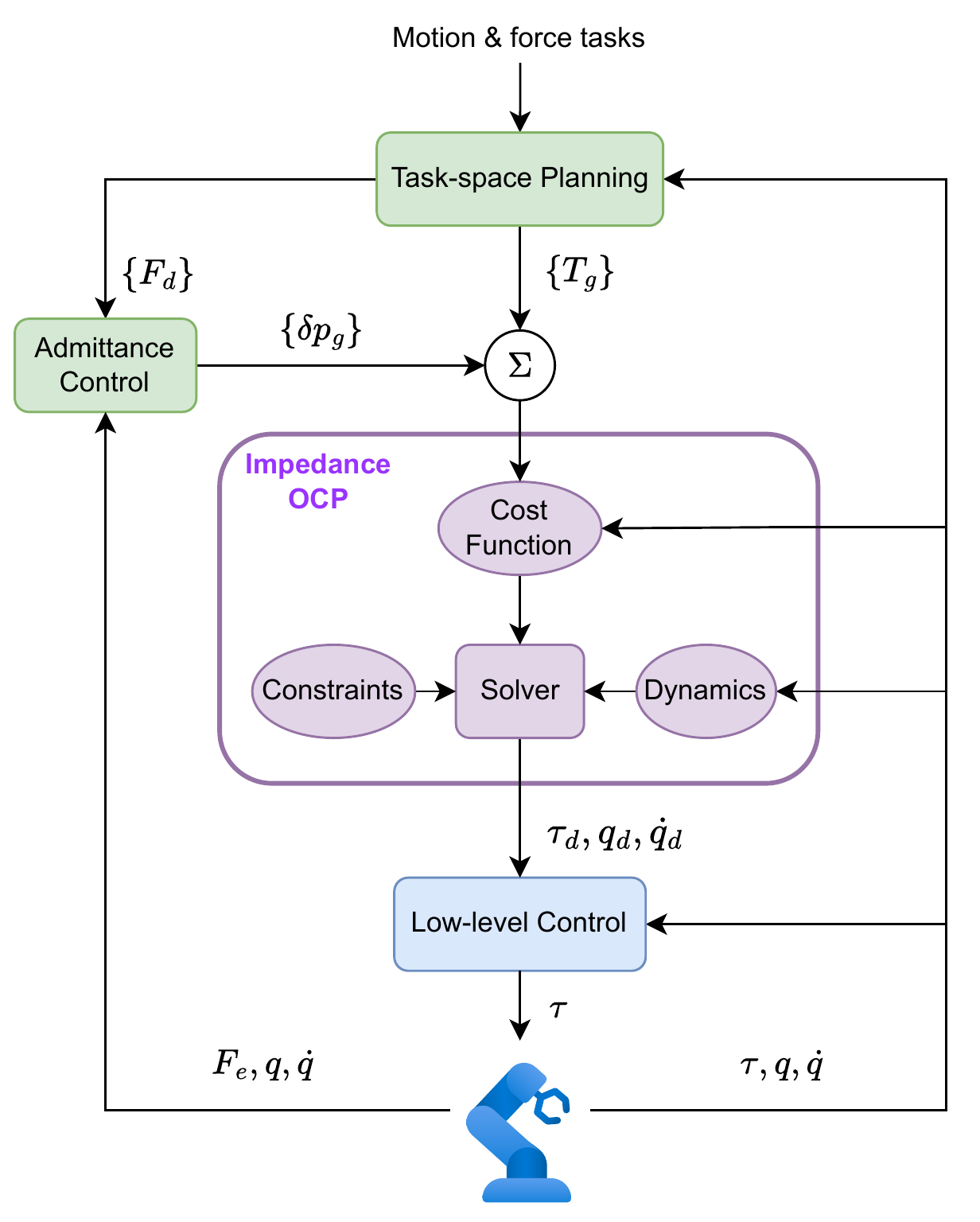}
    \caption{Task-space Impedance-Admittance NMPC Architecture}
    \label{fig:control_architecture}
\end{figure}

Our OCP formulation using end-effector costs allows to solve tasks in two stages, i.e. the task-space motion-force planning and the generic optimal control, resulting in a unified architecture (see Fig. \ref{fig:control_architecture}) that is generalizable to different types of robots in various tasks.
The high-level task-space planning can be solved using any planning or learning-based methods that output end-effector plans, either discrete or continuous.
For example, locomotion only requires a discrete plan for contact placements and timings.
Manipulation tasks such as pick-and-place can be solved with discrete reaching targets while some tasks, e.g. drawing and polishing, dictate that the end-effector keeps a continuous path.
Finally, end-effector plans are inputs to a generic OCP formulation that computes optimal joint and torque trajectories.
Different from many MPC works, we avoid tracking nominal joint trajectories and therefore use nonlinear robot dynamics.
Compared to classical and learning-based methods, NMPC offers hard constraint satisfaction and online planning.

The salient points of our architecture are:
\begin{itemize}
    \item Instead of tracking a nominal trajectory or formulating all task requirements into a single OCP, we rely on online task-space planners to break down manipulation and locomotion tasks into end-effector costs and constraints that define our generic OCP in real time;
    \item The task-space cost formulation also enables to regulate the end-effector impedance stiffness, crucial to improve robustness to unknown environment and contacts;
    \item We close the force-feedback loop by admittance, synthesizing an Impedance-Admittance NMPC (IA-NMPC);
    \item The OCP is solved online to generate optimal joint position, velocity and torque trajectories that are sent directly to the low-level controller, e.g. pure torque control, position control, or with a joint-space PD control.
\end{itemize}

\section{BACKGROUND}

\subsection{Clarifications on dynamics model with contacts}
Denote $q$ the robot configuration and $u$ the control torque.
The dynamics of fixed-base and floating-base robots is:
\begin{equation}\label{eq:dynamics}
    M(q)\ddot{q} + C(q,\dot{q})\dot{q} + \tau_g(q) = S^T u + J_c^T F_c + J_e^T F_{e}
\end{equation}
where $M(q)$ is the robot's inertial matrix; $\tau_g(q)$ is the gravity term, $C(q,\dot{q})\dot{q}$ is Coriolis and centrifugal forces, and $S$ is a selection matrix (for fixed-base robots, $S=I$).
We would like to clarify the distinction between rigid contact wrench $F_c$ and external interaction wrench $F_{e}$. 
Rigid contact is associated with holonomic scleronomic constraint \cite{mastalli2020crocoddyl, peters2005unifying}:
$
    J_c(q)\dot{q} = 0 \Rightarrow J_c(q)\ddot{q} + \dot{J}_c(q)\dot{q} = 0
$
and its contributed torque $J_c^TF_c$ is derived from Gauss's principle of least constraint, where the Lagrangian multiplier is also the contact wrench $F_c$.
In contrast, external interaction wrench $F_{e}$ does not require end-effector constraints.
For example, the gravity term $\tau_g(q)$ is the contribution of gravity applied to the center of mass (COM), which can be written as $\tau_g(q) = J_{COM}^Tmg$ but no constraint is imposed. 
Therefore, $F_e$ may include gravitational forces due to modeling errors.

In this work, we only enforce rigid contact constraint on the standing feet in floating-base robots, while the fixed-base robot's end-effector and floating-base robot's swing feet are not subjected to contact constraints and their force contributions $J_e^T F_{e}$ are not included in our prediction model even when they interact with the environment, i.e.
\begin{equation}\label{eq:prediction}
    M(q)\ddot{q} + C(q,\dot{q})\dot{q} + \tau_g(q) = S^T u + J_c^T F_c
\end{equation}
which is valid for a number of reasons. 
Firstly, contacts may not occur as planned and are generally not rigid contacts, so predicting exact contact timings and force evolution is prone to errors.
Secondly, we will show that the mismatch between our prediction model (\ref{eq:prediction}) and the full dynamics (\ref{eq:dynamics}), together with our cost function, characterize a task-space impedance which can be regulated and is robust to unknown interactions.

\subsection{Optimal Control Problem}
For the purpose of recovering the control law explicitly, in this section, we focus on the unconstrained OCP (\ref{eq:ocp_discrete}a-b).

The optimal trajectories of states $\mathbf{x}=x_0,\dots,x_N$, where $x_k=(q_k,\dot{q}_k)$, and controls $\mathbf{u}=u_0,\dots,u_{N-1}$ must satisfy the Karush–Kuhn–Tucker (KKT) conditions (cf. \cite{jordana2025structure}): 
\begin{gather} \label{eq:kkt}
    \begin{bmatrix}
        \Gamma_1 & M_1^T & 0 & \cdots & 0\\
        M_1 & \Gamma_2 & M_2^T & \cdots & 0\\
        0 & M_2 & \Gamma_3 & \ddots & 0\\
        \vdots & \vdots & \ddots & \ddots & \vdots\\
        0 & 0 & 0 & \ddots & \Gamma_N
    \end{bmatrix}
    \begin{bmatrix} s_1 \\ s_2 \\ s_3 \\ \vdots \\ s_N \end{bmatrix}
    = \begin{bmatrix} g_1 \\ g_2 \\ g_3 \\ \vdots \\ g_N \end{bmatrix} \\
    \Gamma_k = \begin{bmatrix}
        R_{k-1} & 0 & -B_{k-1}^T \\
        0 & Q_k & I\\
        -B_{k-1} & I & 0
    \end{bmatrix},
    M_k^T = \begin{bmatrix}
        0 & 0 & 0\\
        0 & 0 & -A_k^T\\
        0 & 0 & 0
    \end{bmatrix} \nonumber\\
    s_k = \begin{bmatrix} \Delta u_{k-1} \\ \Delta x_k \\ -\lambda_k \end{bmatrix}, \quad
    g_k = \begin{bmatrix} -b_{k-1} \\ -a_k \\ \gamma_k \end{bmatrix} \nonumber
\end{gather}
where $Q_k = \nabla_{xx} l_k$, $R_k = \nabla_{uu} l_k$, $a_k = \nabla_x l_k$, $b_k = \nabla_u l_k$, 
$\gamma_k = f_k-x_{k+1}$, and the dynamics is linearized at the current guess with $A_k = \nabla_x f_k$, $B_k = \nabla_u f_k$.
We do not include $\nabla_{xu}\Box$ terms which are zeros for our cost function (\ref{eq:cost}).

\begin{assumption}(Convergence)
    Regardless of the choice of solvers, we assume convergence after sufficient iterations:
    $$\Delta x_k \rightarrow 0, \quad \Delta u_k \rightarrow 0 \quad \forall k $$
\end{assumption}

Substituting into (\ref{eq:kkt}), we recover the Riccati recursion: starting with $\lambda_N = a_N$ then iteratively computing as follows
\begin{equation}\label{eq:riccati}
\begin{split}
    B_{k-1}^T \lambda_k &= -b_{k-1} \quad \forall k = N,\cdots,1 \\
    -\lambda_k + A_k^T \lambda_{k+1} &= -a_k \quad \forall k = N-1,\cdots,1
\end{split}
\end{equation}

\section{TASK-SPACE IMPEDANCE NMPC}
In this section, we would like to apply the Riccati recursion (\ref{eq:riccati}) to recover the converged solution for an OCP with a generic cost function that will be used in our nonlinear MPC architecture.
In particular, we will show that the optimal control law characterizes task-space impedance control.
This insight will allow us to regulate task-space impedance and to close the force-feedback loop in NMPC in the next section.

\subsection{Cost function}
To benefit from the advantages of nonlinear MPC, we avoid tracking nominal joint trajectories, hence do not use the joint position cost, while joint velocity cost is necessary for smooth trajectories.
Notably, we use the end-effector pose cost, which is widely used to plan optimal trajectories to reach a goal.
Consider a cost function that consists of the end-effector cost, velocity and control regularization costs:
\begin{equation}\label{eq:cost}
    l_k(x_k,u_k) = \frac{1}{2} \|\Delta p_k\|^2_{Q_P} + \frac{1}{2} \|\dot{q}_k\|^2_{Q_V} + \frac{1}{2} \|u_k - u_k^r\|^2_R
\end{equation}
where $u_k^r$ is a reference control, typically the gravity compensation torque.

\begin{definition}\label{def:residual}
We define the end-effector residual:
\begin{equation}\label{eq:rel_pose}
    \Delta p = - T_g \ominus T_e = - \log_6 (T_e^{-1} T_g) \in se(3)
\end{equation}
where $\ominus$ is the right-minus \cite{sola2018micro} and $(-\Delta p)$ is the relative pose (twist) from the end-effector frame to a goal frame, expressed in the end-effector frame.
The sign flip is merely for easier interpretation, e.g. common convention for relative position is $x_e - x_g$.
The task Jacobian is
\begin{equation}
    J_{\Delta p} = \frac{\partial \Delta p}{\partial q} = \text{Jlog}_6(T_g^{-1} T_e) _eJ_e(q)
\end{equation}
where $\text{Jlog}_6(\Box)$ is the right Jacobian of the logarithm \cite{carpentier2019pinocchio}. 
\end{definition}


\subsection{Combined Forward Kinematics-Dynamics}
\begin{definition}\label{def:kin_dyn}
We define the following variables:
\begin{subequations}
\begin{align}\label{eq:new_var}
    y &= y(x) = \begin{bmatrix} \Delta p(q)\\ \dot{q} \end{bmatrix}\\
    w &= w(x,u) = u - u^r(q)
\end{align}
\end{subequations}
and the Forward Kinematics and Forward Dynamics is:
\begin{equation}\label{eq:combined_IK_ID}
    \frac{dy}{dt} =
    \begin{bmatrix}
        J_{\Delta p} \dot{q} \\
        M^{-1} ( S^T w + S^T u^r + J_c^T F_c - \tau_g - C\dot{q} )
    \end{bmatrix}
\end{equation}
\end{definition}

\begin{assumption}\label{ass:grav_comp}
It is common in practice to set the reference control $u^r$ to be the gravity compensation, given that one end of the robot (the fixed base or the standing foot) is in rigid contact with the ground, based on the dynamics model
$$S^T u^r + J_c^T F_c = \tau_g$$
while any modeling errors (e.g., inertial matrices, joint friction, etc.) can be included in the external force $F_e$.
\end{assumption}


Based on the above assumption, we can simplify the combined Forward Kinematics-Dynamics (\ref{eq:combined_IK_ID}):
\begin{equation}
    \dot{y} \approx
    \begin{bmatrix} 0 & J_{\Delta p} \\ 0 & -M^{-1}C \end{bmatrix} y
    + \begin{bmatrix} 0 \\ M^{-1}S^T \end{bmatrix} w
\end{equation}
which can be discretized using the forward Euler method:
\begin{equation}\label{eq:simplified_IK_ID}
\begin{split}
    y_{k+1} &= \begin{bmatrix} I & J_{\Delta p_k} \Delta t \\ 0 & I-M^{-1}C\Delta t \end{bmatrix} y_k 
    + \begin{bmatrix} 0 \\ M_k^{-1}S^T \Delta t \end{bmatrix} w_k \\
    &= Y_k y_k + W_k w_k
\end{split}
\end{equation}

\subsection{Recovering the control law} 
We rewrite the cost function (\ref{eq:cost}) using the new variables:
\begin{equation}\label{eq:cost_yw}
    l_k(y_k,w_k) = \frac{1}{2} \|y_k\|^2_Q + \frac{1}{2} \|w_k\|^2_R,
    \quad Q = \begin{bmatrix} Q_P & 0 \\ 0 & Q_V \end{bmatrix}
\end{equation}

Applying the Riccati recursion (\ref{eq:riccati}) to our cost function (\ref{eq:cost_yw}) and Forward Kinematics-Dynamics (\ref{eq:simplified_IK_ID}), starting at $k=N$:
\begin{align}
    \lambda_N &= (\nabla_x y_N)^T Q y_N 
    = \begin{bmatrix} J_{\Delta p_N}^T & 0\\ 0 & I \end{bmatrix} Q y_N \nonumber\\
    &= \bar{J}^T_N Q y_N = P_N y_N \\
    R w_{N-1} &= -B_{N-1}^T \lambda_N \nonumber\\
    &= -B_{N-1}^T P_N (Y_{N-1} y_{N-1} + W_{N-1} w_{N-1}) \nonumber\\
    w_{N-1} &= - (R + B_{N-1}^T P_N W_{N-1})^{-1} B_{N-1}^T P_N Y_{N-1} y_{N-1} \nonumber\\
    &= - K_N y_{N-1}
\end{align}
then this backward pass can be summarized as follows:
\begin{equation}\label{eq:ts_riccati}
\begin{split}
    \lambda_N &= P_N y_N,\quad P_N = \bar{J}^T_N Q \\
    w_{k-1} &= - K_k y_{k-1} \\
    K_k &=  (R + B_{k-1}^T P_k W_{k-1})^{-1} B_{k-1}^T P_k Y_{k-1} \\
    \lambda_{k-1} &= a_{k-1} + A_{k-1}^T \lambda_k
    = P_{k-1} y_{k-1} \\
    P_{k-1} &= \bar{J}^T_{k-1}Q + A_{k-1}^T P_k Y_{k-1} - A_{k-1}^T P_k W_{k-1} K_k
\end{split}
\end{equation}
Finally, MPC applies the first control to the robot:
\begin{equation}\label{eq:law}
    w_0 = - K y_0 \quad
    \text{or}\quad u_0 - u^r_0 = - K \begin{bmatrix} \Delta p_0 \\ \dot{q}_0 \end{bmatrix}
\end{equation}
where $K=K_1$ is called the Riccati gain.

\subsection{Task-space Impedance of our OCP formulation}
Since $S^T u^r + J_c^T F_c \approx \tau_g$ if $u^r$ is the gravity compensation torque (Assumption \ref{ass:grav_comp}), we can rewrite the full dynamics (\ref{eq:dynamics}):
\begin{equation}\label{eq:compact_dynamics}
    M\ddot{q} + C\dot{q} = S^T w + J_e^T F_{e}
\end{equation}
We left-multiply both sides by $J_eM^{-1}$ and rearrange to get
\begin{equation}
    F_{e} = (J_e M^{-1} J_e^T)^{-1}
    \left( J_e\ddot{q} + J_e M^{-1} C\dot{q} - J_e M^{-1} S^T w \right) \nonumber
\end{equation}
and since
$\ddot{p} = J_e\ddot{q} + \dot{J}_e\dot{q}
\Longrightarrow
J_e\ddot{q} = \ddot{p} - \dot{J}_e\dot{q}$,
we recover the task-space dynamics of the end-effector (cf. \cite{khatib1987unified}):
\begin{align}\label{eq:ts_dyn}
    F_{e} + J^\dagger S^T w = \Lambda \ddot{p} - \Lambda \dot{J}_e\dot{q} + J^\dagger C\dot{q}
\end{align}
where $\Lambda = (J_e M^{-1} J_e^T)^{-1}$ is the task-space inertial matrix, $J^\dagger = (J_e M^{-1} J_e^T)^{-1} J_e M^{-1}$ is the generalized pseudo-inverse of the end-effector's Jacobian, i.e. $J^\dagger J_e^T=I$.
It is worth highlighting that $J^\dagger$ is not used for Inverse Kinematics anywhere in our proof.
Using the optimal control $w_0$ from (\ref{eq:law}), the gravity-compensated control wrench is 
\begin{equation}\label{eq:wrench}
    F \triangleq J^\dagger S^T w_0 = -J^\dagger S^T K y
\end{equation}
where we remove the subscripts in $y_0, \Delta p_0, q_0$ for readability.

Finally, denote $\eta(q,\dot{q})=(J^\dagger C - \Lambda\dot{J}_e)$, split $Ky = K_P \Delta p + K_D \dot{q}$ and substitute (\ref{eq:wrench}) into (\ref{eq:ts_dyn}), we obtain a closed-loop dynamics that obeys task-space impedance law:
\begin{equation}\label{eq:impedance}
    F_{e} = \Lambda \ddot{p} + (\eta+J^\dagger S^T K_D) \dot{q}
    + J^\dagger S^T K_P \Delta p
\end{equation}
where the impedance stiffness and damping $K_P, K_D$ can be scaled through the weight matrices $Q_P, Q_V$ as they back-propagate through the Riccati recursion (\ref{eq:ts_riccati}).

This task-space impedance law attracts the end-effector to a steady state by the following equilibrium:
\begin{equation} \label{eq:ss_wrench}
    F_{e,\infty} = J^\dagger S^T K_P  \Delta p_{\infty}
\end{equation}
which can prevent the end-effector from deviating away under unmodeled forces $F_e$.
Next, we shall analyze the contraction and passivity of this Impedance NMPC (I-NMPC).



\section{MOTION-FORCE NMPC}

\begin{figure}
    \centering
    \includegraphics[width=0.6\linewidth]{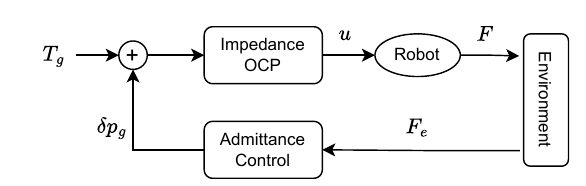}
    \caption{Simplified diagram of Impedance-Admittance NMPC}
    \label{fig:imp_adm}
\end{figure}

\subsection{Partial Contraction and Differential Passivity}
Since our system is nonlinear and time-varying, we follow the partial contraction analysis framework \cite{wang2005partial}. 
Consider a virtual system for our closed-loop dynamics (by substituting (\ref{eq:wrench}) into (\ref{eq:compact_dynamics})) with respect to a virtual state $z=\dot{q}$, the external forces $F_e$, and the relative twist of end-effector goal $v_{eg} = -\Delta p$ (recall Definition \ref{def:residual}), the variation dynamics is:
\begin{equation}\label{eq:virtual}
    M\delta\dot{z} + C\delta z = - S^T (K_P \delta\Delta p + K_D \delta z) + J_e^T \delta F_{e}
\end{equation}

We define the following differential Lyapunov function which is also differential storage function:
\begin{equation}
    \delta V = \frac{1}{2} \delta\dot{q}^T M(q) \delta\dot{q} = \frac{1}{2} \delta z^T M \delta z
\end{equation}
Noting that $\delta z^T (\dot{M}-2C) \delta z=0$ and $\delta \dot{p}_e = J_e \delta z$, we obtain:
\begin{align}\label{eq:passivity}
    \delta\dot{V} &= \delta z^T (S^T K_P \delta v_{eg} - S^T K_D \delta z + J_e^T \delta F_e) \\
    &= \delta \dot{p}_e^T \delta  F_e + \delta z^T S^T K_P \delta v_{eg} - \delta z^T S^T K_D \delta z \nonumber
\end{align}

\subsubsection{Partial contraction}
Without the moving goal and external force variation, (\ref{eq:passivity}) simplifies to 
$\delta\dot{V} = - \delta z S^T K_D \delta z$.
Condition: if $S^TK_D \geq \alpha I, \alpha>0$, the system is partially contracting, guaranteeing differential stability and convergence, in our case, to the task-space equilibrium (\ref{eq:ss_wrench}).
Compared to standard contraction which guarantees convergence to a single trajectory, our system allows freedom in the joint space which is crucial for optimal redundancy exploitation of manipulators and adapting behaviors of legged robots.

\subsubsection{Differential passivity}
Taken the moving goal and external force variations into account, the differential passivity condition $\delta\dot{V} \leq \delta \dot{p}_e^T \delta  F_e + \delta z^T S^T K_P \delta v_{eg}$ is satisfied if $S^TK_D \geq 0$.
If $S^TK_D > 0$, the system is \textit{strictly differentially passive}, which means it always dissipates energy faster than the energy transfer rate from the external forces $\delta\dot{p}_e^T \delta F_e$ and from the virtual energy created by the moving goal $\delta z^T S^T K_P \delta v_{eg}$.
Given this passivity, we can safely close the force-feedback loop to regulate external wrench in NMPC by adjusting the end-effector's goal pose using admittance law (Fig. \ref{fig:imp_adm}), provided that an interconnection of admittance control with passive systems is passive \cite{kim2016passivity}.

In summary, the condition to obtain both partial contraction and differential passivity is: $S^TK_D \geq \alpha I, \alpha>0$ which is achievable in practice by tuning the cost weight matrix $Q$, e.g. high enough to dominate during back propagation (\ref{eq:ts_riccati}).

\subsection{Impedance-Admittance NMPC}
To track a desired force $F_d$ and motion $p_d$, we propose an admittance step that adapt the relative end-effector pose $\Delta p$ by updating the goal frame's transformation matrix $T_g$ in (\ref{eq:rel_pose}) by a twist $\delta p_g$ by integrating a desired dynamics: 
\begin{equation}\label{eq:admittance}
    M_a (\ddot{p}_g-\ddot{p}_d) + D_a (\dot{p}_g-\dot{p}_d) + K_a (p_g-p_d) = F_e - F_d
\end{equation}
where $M_a, D_a, K_a$ are positive-diagonal matrices that represent the desired inertia, damping and stiffness; $p_d,\dot{p}_d,\ddot{p}_d$ are the desired 6D pose, velocity and acceleration with respect to a fixed frame, i.e. $p=[r,\Theta]$, $r=[x,y,z]$, $\Theta=[\phi,\theta,\psi]$.

\subsubsection{Staying mode}
$p_d$ is fixed, and $\ddot{p}_d, \dot{p}_d, F_d=0$ so that the object returns to a desired equilibrium $p_e=p_d, F_e=0$.
\subsubsection{Floating mode}
$p_d=p_g$, and $\ddot{p}_d, \dot{p}_d, F_d=0$ so that the object can be moved to any poses in space.
\subsubsection{Force tracking}
setting $p_d=p_g$, $\ddot{p}_d=\ddot{p}_g$, $\dot{p}_d=0$, and $F_d \neq 0$ simplifies (\ref{eq:admittance}) to $D_a \dot{p}_g = F_e-F_d$ which integrates into proportional force control: $\delta p_g = K_f (F_e-F_d)$.
One could also include a derivative term to obtain PD force control, i.e. 
$\delta p_g = K_{fP} (F_e-F_d) + K_{fD}(\dot{F}_e-\dot{F}_d)$.

Since the OCP outputs the optimal joint and torque trajectories, our framework can run on either position-controlled or torque-controlled robots.
In position-control mode, we obtain Admittance NMPC (A-NMPC), and in torque-control mode, Impedance-Admittance NMPC (IA-NMPC).

\section{EXPERIMENTS}
We perform all experiments using a laptop with CPU AMD 8940HX.
We test two OCP solvers, FDDP \cite{mastalli2020crocoddyl} and \textit{mim\_solvers} \cite{jordana2025structure}, and both generate compliant behaviors since the task-space impedance is the underlying control law of the OCP and agnostic to numerical methods (cf. section IV).
In the following experiments, OCPs are solved by \textit{mim\_solvers} which treats inequality requirements as hard constraints.

\subsection{Experiment 1: Task-space Impedance in manipulation}

\begin{figure}[t]
    \centering
    \includegraphics[width=0.32\linewidth]{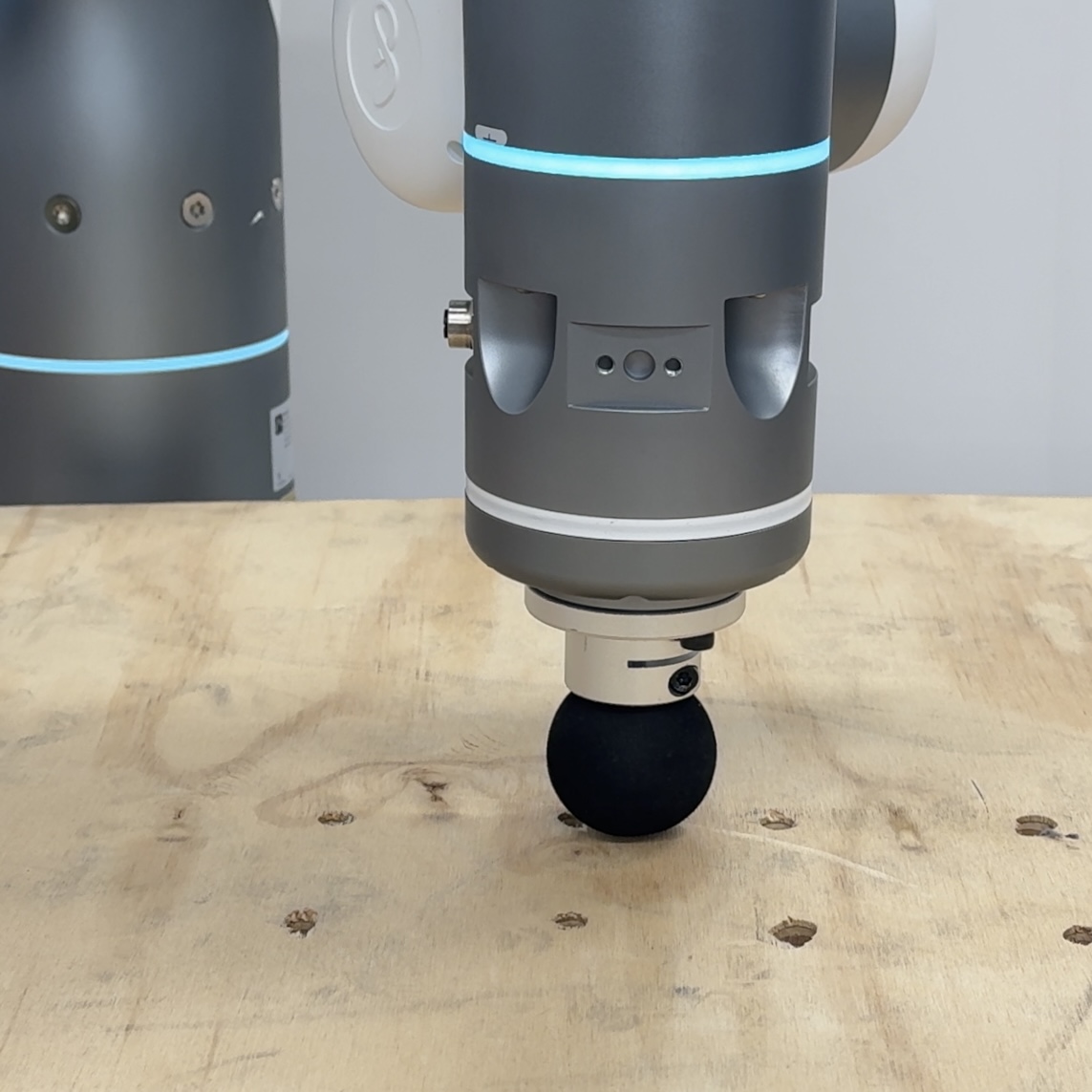}
    \includegraphics[width=0.32\linewidth]{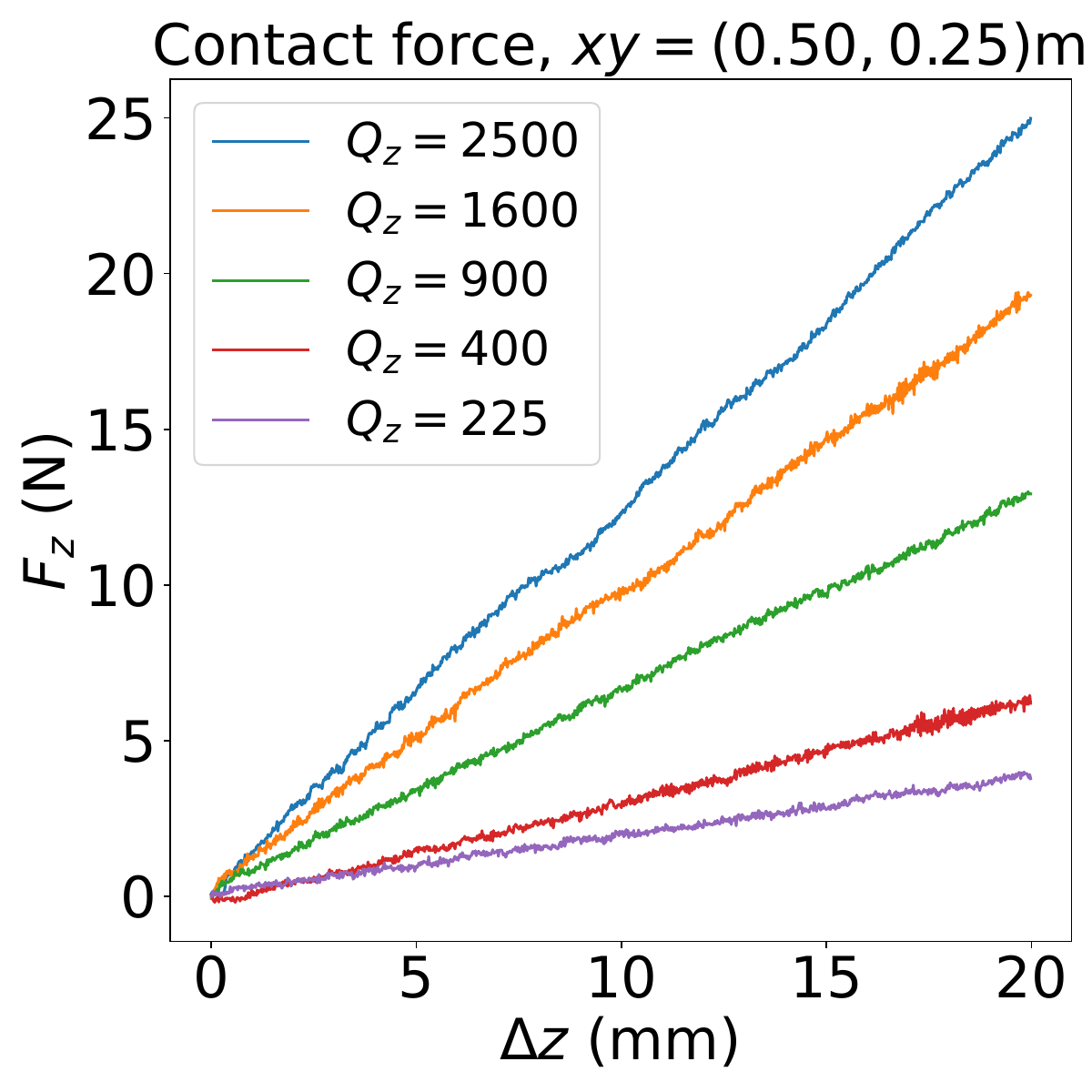}
    \includegraphics[width=0.32\linewidth]{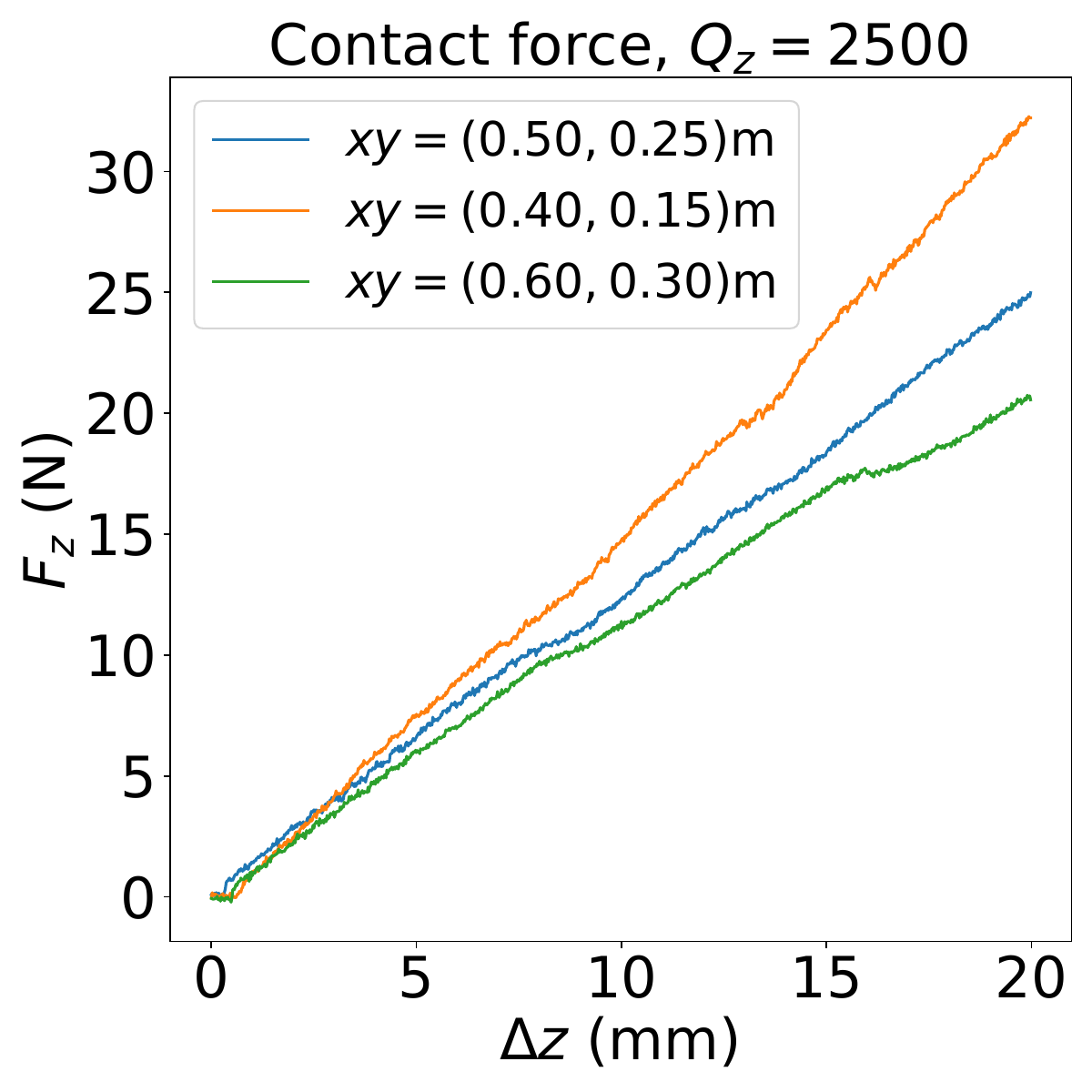}
    \caption{Press test: contact forces when changing the relative goal pose $\Delta z$, the $z$ component of weight matrix $Q$, and the end-effector's position.}
    \label{fig:press}
\end{figure}

In section IV, we characterized the relationship between the task-space impedance and the end-effector costs.
The task-space compliance has been observed in a recent work \cite{kleff2021high} but was credited to high-frequency closed-loop MPC.

To validate our theoretical analysis of the task-space impedance as a result of the end-effector cost, we set up a pressing test as follows.
A 7-DOF manipulator (Flexiv) presses at different locations on a wooden surface (Fig. \ref{fig:press}) with different weight matrices for the end-effector cost.
A force/torque sensor measures contact force changes at the end-effector. 
The first plot shows vertical contact force increases linearly as the goal pose is set further below the surface.
It also shows that as the $z$ component of the weight matrix increases, the stiffness (the slope) is scaled up.
This stiffness is state-dependent (second plot).
These results validate force-position relationship characterized by the impedance law (\ref{eq:ss_wrench}).
This observation suggests to control contact forces by adapting the end-effector goal pose (Admittance), while the stiffness of the end-effectors can be adjusted by the cost weight matrix (Impedance).



\begin{figure}[t]
    \centering
    \includegraphics[width=0.42\linewidth]{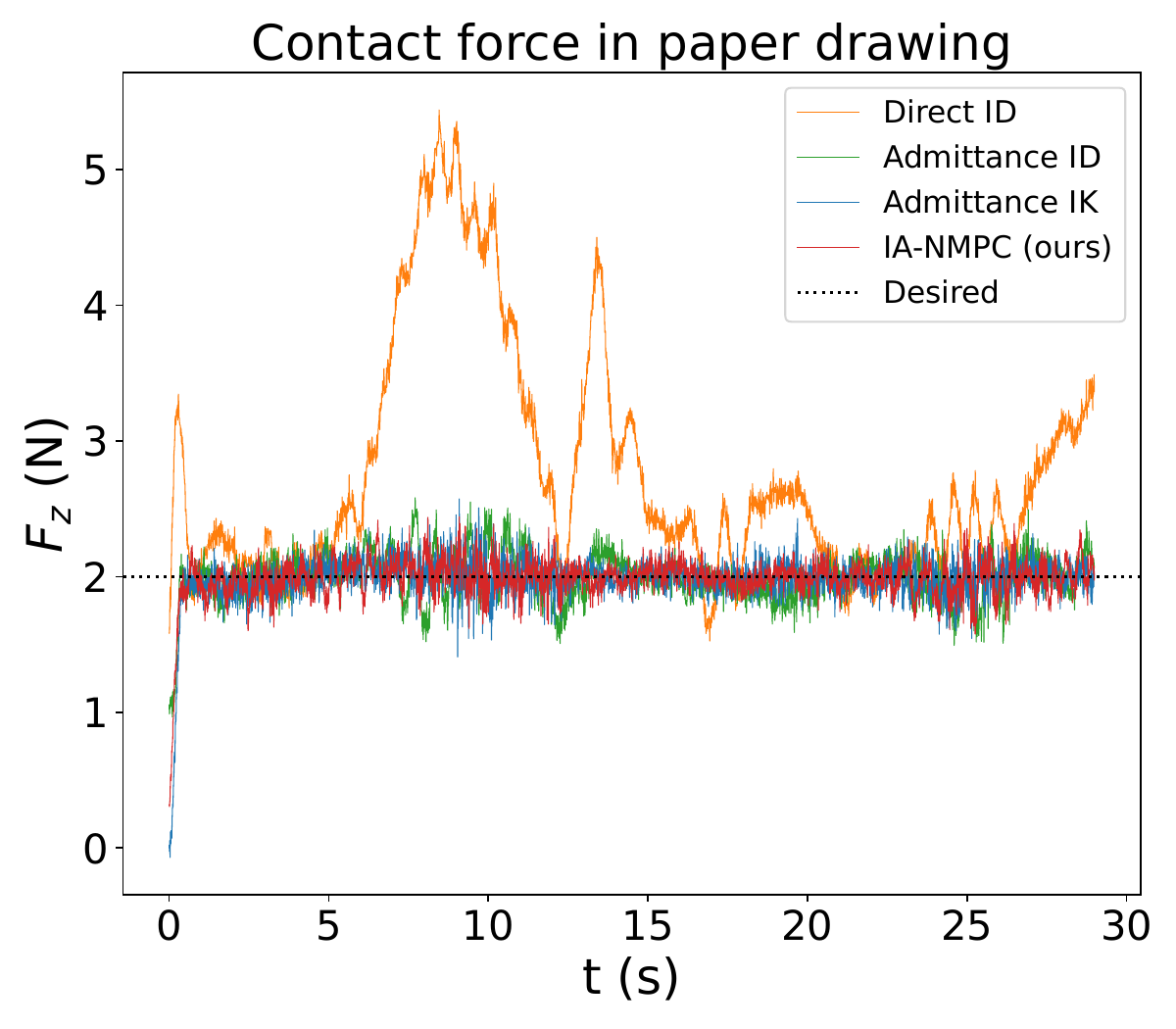}
    \includegraphics[width=0.42\linewidth]{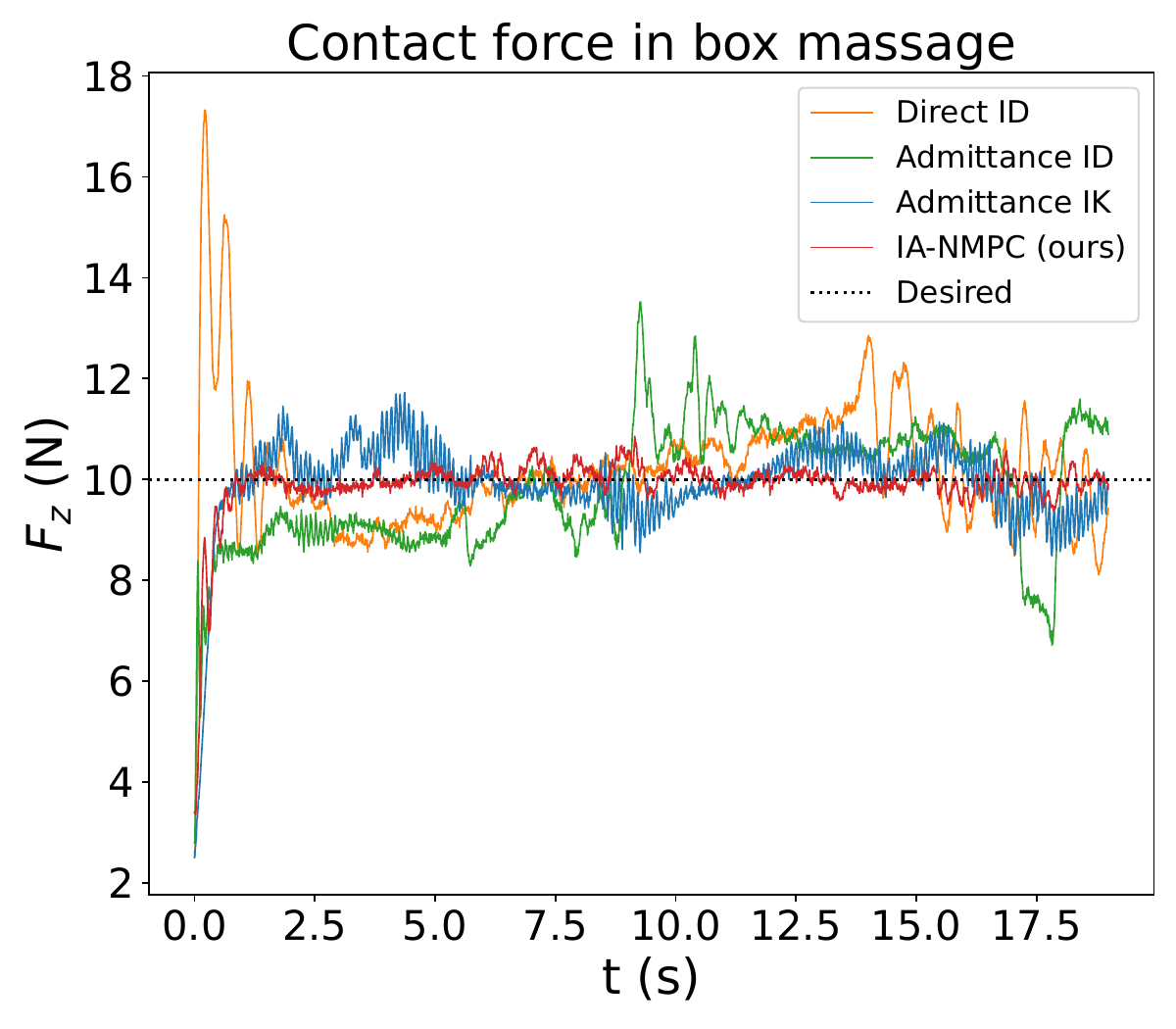}
    \caption{Comparison of force tracking precision between different methods.}
    \label{fig:force}
\end{figure}

\subsection{Experiment 2: Drawing on flat and deformable surfaces}
In this series of experiments (see Fig. \ref{fig:highlights}a-b), Flexiv must draw with $1.0N$ on a white board, $2.0N$ on paper, and $0.5N$ on the balloon.
Similar tasks such as polishing/massage (Fig. \ref{fig:highlights}c) can be done by setting the desired contact force, e.g. $10N$.
For these tasks, one could use the force tracking admittance control in $z$ direction: $\delta z_g = k_{fp} (f_{z,e}-f_{z,d}) + k_{fd} \dot{f}_{z,e}$ where the admittance coefficients $k_{fp}, k_{fd}$ are tuned for each material and method to the best of our effort:
\begin{itemize}
    \item Massage: $k_{fp}=0.005m/N, k_{fd}=0.0005$ for all.
    \item Paper drawing: $k_{fp}=0.00007m/N, k_{fd}=0$ for all.
    \item Board drawing: we re-tune $k_{fp}=0.00003$ except for our IA-NMPC where we keep old value $k_{fp}=0.00007m/N$ but soften the impedance by lowering $Q_z$.
    \item Balloon: similarly, NMPC uses $k_{fp}=0.002m/N$ with low $Q_z$ while others are tuned to $k_{fp}=0.0005m/N$.
\end{itemize}

In board and balloon drawing, our IA-NMPC works well with the same admittance coefficients as others but we want to test different admittance coefficients to show the benefit of the impedance layer in changing the contact stiffness.

We use raw force/torque sensor data without any additional estimator/filter.
All drawing experiments use the same MPC setup: update frequency $100\textit{Hz}$, receding horizon $N=20\ \text{nodes} \times 0.01s$.
Keep all other parameters the same, we compare our method IA-NMPC (pure torque mode) and A-NMPC (position mode) with three baseline controllers: 
\subsubsection{Classical admittance control (Admittance IK)}: passing desired end-effector velocity adjusted by the admittance law to a QP solver to compute joint velocities.
\subsubsection{Torque-control admittance (Admittance ID)} adding another layer to Admittance IK that maps desired joint acceleration to torque via Inverse Dynamics.
\subsubsection{Direct motion-force control} using Inverse Dynamics for motion-tracking directions and null-space force control assuming rigid contact in force-tracking direction.

Classical admittance control is already the state-of-the-art, so we do not compare with direct force-control MPC since these methods require modifying the dynamics model and optimization variables.
Since our robot model is not accurate, the torque-control baselines must be coupled with a low-level PD controller.
Meanwhile, our IA-NMPC sends pure torque commands to the robot, validating the task-space impedance without the help of the joint-space PD impedance law.

\begin{figure}[t]
    \centering
    \includegraphics[width=0.38\linewidth]{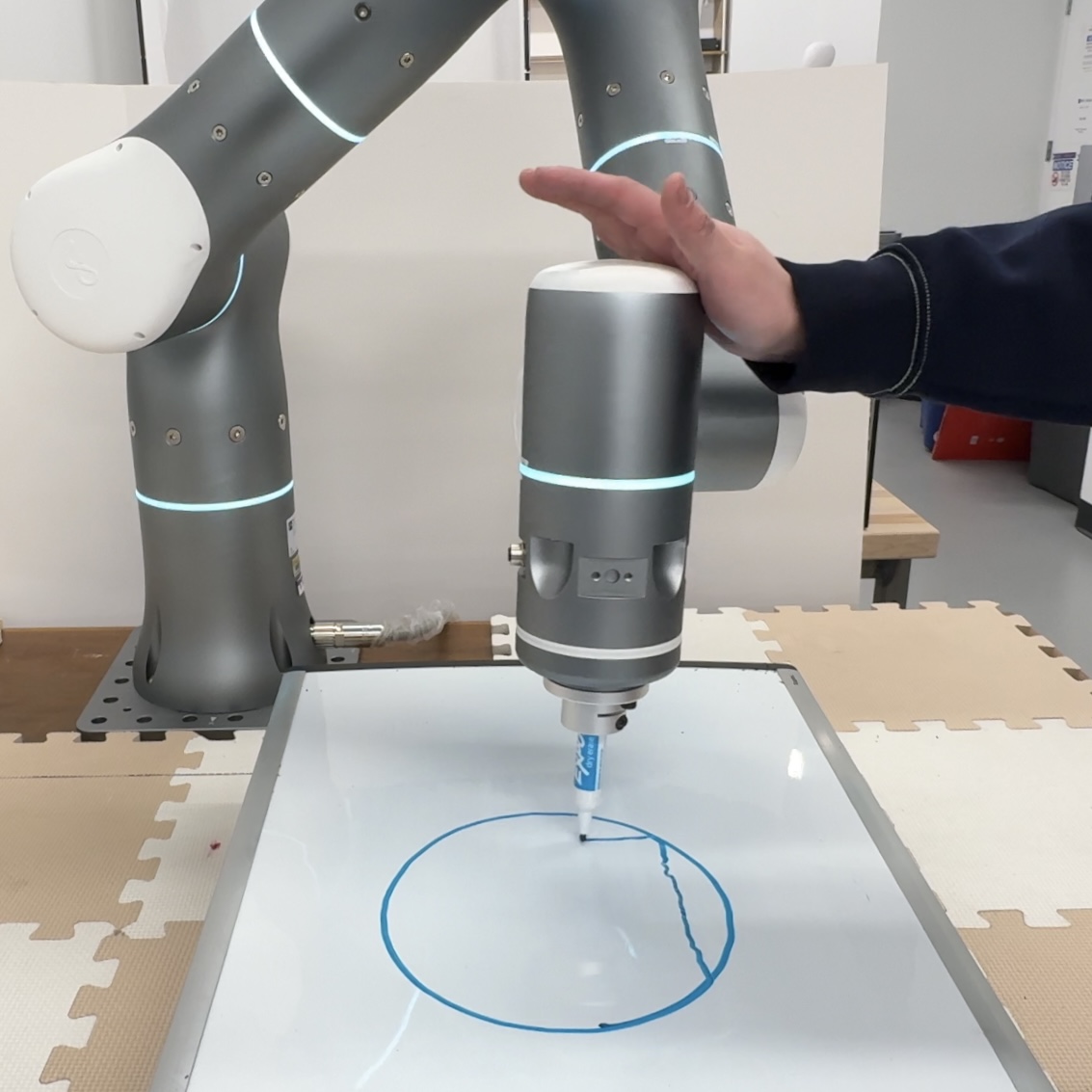}
    \includegraphics[width=0.38\linewidth]{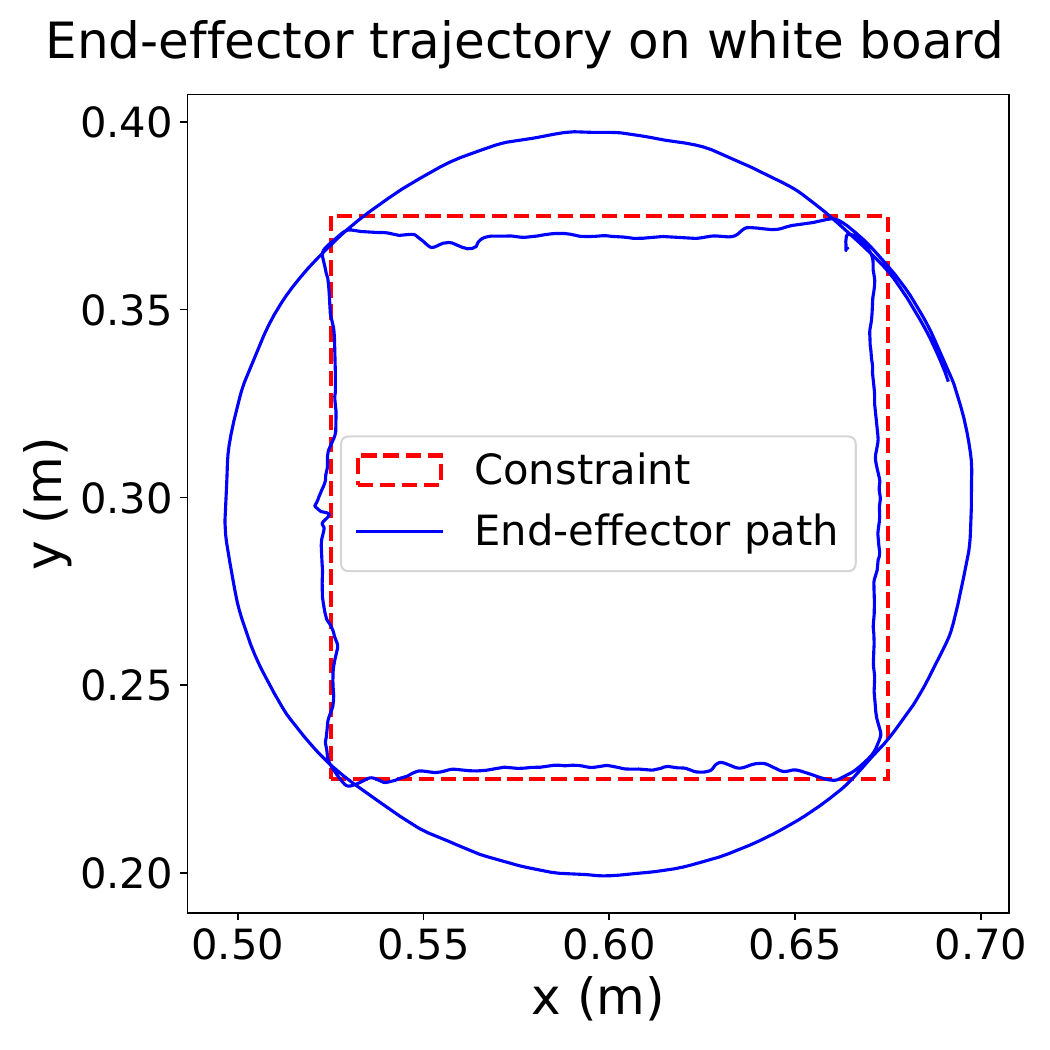}
    \caption{Drawing under disturbances with real-time constraint addition.}
    \label{fig:constraint}
\end{figure}

\begin{table*}[t]
\caption{Contact force and position tracking precision on different surfaces and desired force:\\
$\bigcirc$ Board (1.0N) -- $\heartsuit$ Paper (2.0N) -- $\heartsuit$ Balloon (0.5N) -- $\infty$ Box massage (10.0N)}
\centering
\begin{tabular}{c | c c c c c}
\hline
Method  & IA-NMPC (ours)            & A-NMPC (ours)             & Admittance IK             & Admittance ID              & Direct Force Control       \\
(Mode)  & (Pure Torque)                  & (Position)                & (Position)                & (Torque + PD)              & (Torque + PD)              \\
\hline
Board   & $0.98\pm0.23$N (0.65mm)   & $0.98\pm0.27$N (0.05mm)   & $0.98\pm0.15$N (0.02mm)   & $0.99\pm0.25$N (6.86mm)    & $1.11\pm0.29$N (6.09mm)    \\
Paper   & $2.00\pm0.12$N (0.87mm)   & $2.00\pm0.12$N (0.17mm)   & $2.00\pm0.12$N (0.16mm)   & $2.00\pm0.16$N (6.11mm)    & $2.81\pm0.87$N (5.67mm)    \\
Balloon & $0.50\pm0.08$N (1.48mm)   & $0.50\pm0.09$N (0.37mm)   & $0.50\pm0.12$N (0.11mm)   & $0.49\pm0.10$N (6.42mm)    & $2.06\pm1.08$N (8.80mm)    \\
Massage & $10.0\pm0.2$N (0.87mm)    & $10.0\pm0.2$N (0.43mm)    & $10.0\pm0.5$N (0.24mm)    & $10.0\pm1.1$N (11.9mm)     & $10.1\pm0.9$N (7.16mm)     \\
\hline
\end{tabular}
\label{tab:draw}
\end{table*}

Results are shown in Table \ref{tab:draw} and Fig. \ref{fig:force}.
Overall, IA-NMPC, A-NMPC and Admittance IK keep highly precise contact forces throughout all experiments.
Notably, these methods enable the robot to draw with $0.50$ on a balloon with standard deviation of $0.08-0.12N$ without knowledge of the balloon's curvature and deformations.
In term of position accuracy, classical Admittance IK and A-NMPC achieved the smallest deviation since they are position-controlled.
Meanwhile, IA-NMPC achieved slightly lower accuracy due to its compliance to friction in other directions.
Nonetheless, a standard deviation of $0.65-1.48mm$ is highly accurate for most cases and the impedance stiffness can be adjusted when extreme position accuracy is needed by the applications.

Torque-control Admittance ID, though achieves good force tracking performance, cannot plan when it is deviated from the reference due to friction, leading to poor motion tracking in massage task.
In contrast, direct force control performs poorly in drawing tasks since it depends on a flat surface model, and improves in the massage where the rigid contact assumption becomes more relevant under high contact forces.

The advantage of our method over Admittance IK is more significant in the massage/polishing task when the desired force of $10N$ making the contact dynamics stiffer which is known to be where Admittance IK performance drops.
In comparison, classical admittance control relies solely on actual movement of the end-effector (infinite stiffness) while our IA-NMPC adds an impedance layer that is equivalent to adding springs in series to soften the contact dynamics.

In addition, MPC framework allows adding and solving hard constraints in real-time, even when subject to human interaction, as shown in Fig. \ref{fig:constraint}.
Together with the adjustable stiffness of the task-space impedance layer as discussed above, our IA-NMPC provides safe and precise interactions which is a critical property in many applications, especially with human-robot interaction as shown in the video.

\subsection{Experiment 3: $SE(3)$ admittance with a heavy object}
We demonstrate a pipeline to manipulate unknown heavy objects in $SE(3)$ (Fig. \ref{fig:highlights}c) without object model in the OCP, which will be useful for human-robot collaboration tasks.

\subsubsection{Estimation of object's inertia properties}
In general, one could estimate the object's inertia matrix using Newton-Euler equations of motion (EOM), such as in \cite{atkeson1985rigid}.
However, since knowledge of the object mass and COM location are sufficient for most cases, we could use the static EOM:
\begin{equation} \label{eq:eom}
    F_e = \begin{bmatrix} f_e\\ \tau_e \end{bmatrix}
    = \begin{bmatrix} m_{o} g\\ c_o \times m_{o} g \end{bmatrix}
\end{equation}

Without the object model, we can still rely on a stiff Impedance NMPC to rotate the end-effector and collect force/torque data at different orientations (see the video).
Object mass $m_{o}$ is estimated by averaging, while the COM $c_o$ is estimated by least square method on the EOM (\ref{eq:eom}).

We can apply this method to estimate inertia properties of every tool used in previous experiments.
For this experiment we attached a $3.48kg$ weight plate to the robot's flange using a coupling mechanism weighted $0.42kg$ (see Fig. \ref{fig:highlights}c). 
Our method estimated the total mass is $3.93kg$ and the COM at $[0.00,0.00,0.04]m$ in end-effector's frame.


\subsubsection{Estimation of external interaction's wrench}
In general, given the object mass and COM obtained in step 1, we can extract any external wrench $[f_{ext}, \tau_{ext}]$ applied to the object.
For a single grasping configuration and slow rotation, external wrench can be determined by:
\begin{equation}
    f_{ext} \approx f_e - m_{o} g, \quad
    \tau_{ext} \approx \tau_e - c_o \times m_{o} g
\end{equation}

\subsubsection{Object manipulation in $SE(3)$}
We implement the $SE(3)$ admittance in (\ref{eq:admittance}) in two modes, i.e. staying and floating, as described in V.B.
Although the robot model is not updated with the object's inertia, we can achieve $SE(3)$ manipulation while the object is being held at the equilibrium according to (\ref{eq:ss_wrench}), which shows the benefit of the impedance-admittance combination.
Moreover, our NMPC framework enables real-time predictive planning under whole-body interactions, which is an improvement to classical methods.

\subsection{Experiment 4: Compliant locomotion for a quadruped in an unknown, unsteady environment with limited sensing}
In this experiment, a quadruped Go2 must walk through unknown, unsteady obstacles (Fig. \ref{fig:highlights}d) without terrain model or visual perception.
Importantly, to stress test the system and validate the task-space impedance, we remove contact detection and translational position/velocity odometry feedback from the controller (floating-base position is set to $[0,0,0]$ and velocity to the commanded velocity $[V_x, V_y, 0]$). We only use joint and gyroscope feedback. The task is therefore quite difficult due to limited feedback. 
There are instances when the obstacles are moving unexpectedly (see Fig. \ref{fig:highlights}d and the video), the robot is still able to compliantly adapt and robustly compute optimal trajectories to recover balance.

Our I-NMPC updates at $33\textit{Hz}$, solving a constrained OCP with joint state and torque bound constraints for a horizon of $10$ to $12\ \text{nodes}\times0.03s$.
The \textit{mim\_solvers} \cite{jordana2025structure} spends less than $10ms$ to converge.
Since Go2 utilizes a low-level PD controller, we send a bundle of the optimal torque, joint position and velocity with $K_{jP}=50.0Nm/rad \times I_{12\times12}$ and $K_{jD}=5.0 Nms/rad\times I_{12\times12}$.

\subsection{Experiment 5: Quadruped Go2 carries human on a chair}
The advantage of our method is that the walking gait can be customized for different purposes by simply changing hyper-parameters, e.g. step height, step duration; and the robot's morphology can adapt to unknown terrains thanks to the impedance at its feet and COM.

Based on the same controller in experiment 4, we simply increase the weight matrix of the COM task cost and set the desired COM further ahead.
With those few parameter changes, the robot is able to pull a chair with human sitting on it (Fig. \ref{fig:highlights}e).
This experiment not only validates the task-space impedance resulting from our cost function design, but also shows the ability to customize robot's behaviors in our framework without any re-formulation or re-training.
In comparison, a recent work \cite{zhang2025sac} proposes a teacher-student RL framework that achieves a similar compliant behavior by training a teacher policy with an explicit compliant law and privileged external force information which was then removed from the student policy training.
We achieve a similar result using passive impedance without force feedback.

\section{CONCLUSIONS}

In this paper, we proposed a generic NMPC framework that solves motion-force requirements in the task space and generates optimal joint trajectories to control the robot in real-time with an intrinsic task-space impedance.
Based on our theoretical analysis of the task-space impedance of a generic OCP formulation with end-effector cost features, we are able to remove unnecessary rigid contact constraints and instead regulate contact forces by adapting the weight matrices and goal poses of the end-effector costs. 
This results in a versatile way to reason about contacts with differential stability and passivity guaranties.
We validated our method through compliant locomotion tasks with a blind quadruped carrying unknown weights in unknown environment and hybrid motion-force control tasks with a redundant manipulator under constraints and unforeseen human-robot interactions.

Within the scope of this paper, the impedance control law is analyzed for the unconstrained OCP.
It would be interesting to study the effect of active constraints that may occur such as joint limits, collision avoidance, friction cone, etc.
These topics involve additional analysis into designing or learning terminal cost functions and constraints capable of approximating long-horizon value function and reachability, which we leave for future investigations.
In addition, our architecture can generate in real-time smooth robot trajectories for any task-and-motion-planning (TAMP) planners or learning policies that output task-space targets, while adding constraint satisfaction guarantee, compliance and force-feedback capabilities.
In practice, the main limitation of our method is that the computation time grows as the DOFs and horizon increase, which introduces delay and limits control frequency.
For repetitive tasks such as walking, our framework could be used as an expert demonstrator to train behavior cloning policies in simulation, thus replacing online computation by offline memorization.

\addtolength{\textheight}{-0cm}   







\bibliographystyle{IEEEtran}
\bibliography{IEEEabrv, references}

\end{document}